%% file: main.tex
\documentclass[10pt,twocolumn,letterpaper]{article}

\usepackage{cvpr}              
\input{preamble}
\definecolor{cvprblue}{rgb}{0.21,0.49,0.74}
\usepackage[pagebackref,breaklinks,colorlinks,allcolors=cvprblue]{hyperref}

\def\paperID{*****} 
\def\confName{CVPR}
\def\confYear{2026}

\title{\vspace{-0.4in}Temporal Evidence Routing with Structured Visual Evidence for TimeLogicQA}

\author{
Yuyang Sun\textsuperscript{1},
Yongliang Wu\textsuperscript{1},
Xingyu Zhu\textsuperscript{2},
Yuxia Chen\textsuperscript{3}, \\
Zhenxiang Jiang\textsuperscript{4},
Yangguang Ji\textsuperscript{4},
Wenbo Zhu\textsuperscript{4},
Yanxi Shi\textsuperscript{4},
Jay Wu\textsuperscript{4},
Shuo Wang\textsuperscript{5},
Xu Yang\textsuperscript{1} \\
\textsuperscript{1}Southeast University
\textsuperscript{2}National University of Singapore
\textsuperscript{3}Independent Researcher \\
\textsuperscript{4}Opus AI Research
\textsuperscript{5}University of Science and Technology of China‌
}

\begin{document}
\maketitle
\input{sec/0_abstract}    
\input{sec/1_introduction}
\input{sec/2_method}
\input{sec/3_exp}
\input{sec/4_conclusion}
{
    \small
    \bibliographystyle{ieeenat_fullname}
    \bibliography{main}
}

\end{document}

%% file: sec/0_abstract.tex
\begin{abstract}
TimeLogicQA evaluates whether video question answering systems can reason over temporal relations such as event existence, ordering, persistence, boundary conditions, and overlap. We address this task with a visual evidence routing pipeline that separates perception from symbolic temporal reasoning. The system first parses each question into event targets, answer mode, candidate options, and temporal operators. It then routes videos according to duration and operator difficulty, using ordered full-frame evidence for short clips and event-focused candidate windows for long videos. A multimodal large language model produces structured visual evidence for the relevant events, while programmatic verifiers recover dense action intervals and a deterministic reducer applies operator-specific temporal rules to produce the final answer. Conservative fusion accepts an answer only when the visual evidence, temporal program, and confidence checks agree, reducing noisy answer flips. On the official test evaluation, our final system achieves an AvgAcc of 81.8.
\end{abstract}

%% file: sec/1_introduction.tex
\section{Introduction}

Video question answering requires a model to combine visual perception with language-conditioned reasoning over time. TimeLogicQA makes this requirement explicit by asking questions that target temporal logic, including whether events occur, whether one event happens before another, whether a relation holds for every occurrence, whether an action continues until a boundary event, and whether two events overlap \cite{swetha2025timelogic}. These questions are difficult for direct video-to-answer prompting because a model must identify the relevant events, preserve their temporal order, and avoid replacing temporal evidence with language priors.

The test split contains 3,000 questions. The answer space is either Boolean or four-way multiple choice. The main temporal operators include basic event existence, before/after relations, always-before constraints, until-style boundary reasoning, and overlap or disjointness checks. Table~\ref{tab:data} summarizes the answer-mode and temporal-operator distribution used by our system routing.

\begin{table}[t]
\centering
\caption{Test-set answer-mode and operator distribution used for routing.}
\label{tab:data}
\footnotesize
\setlength{\tabcolsep}{3pt}
\begin{tabular}{@{}p{0.30\linewidth}p{0.50\linewidth}r@{}}
\toprule
Category & Type & Count \\
\midrule
Answer mode & Boolean & 1,122 \\
Answer mode & Multiple choice & 1,878 \\
\midrule
Temporal operator & Before / after & 1,022 \\
Temporal operator & Always & 961 \\
Temporal operator & Generic / existence / implication & 506 \\
Temporal operator & Overlap / disjoint & 326 \\
Temporal operator & Until & 185 \\
\bottomrule
\end{tabular}
\end{table}

The videos vary substantially in duration and temporal density. Short clips can contain the full reasoning signal in only a few seconds, so losing even a few frames can change the answer. Longer videos make full-video prompting inefficient and can distract the model from the events named in the question. Our method uses different evidence construction strategies for these duration regimes while keeping the same temporal reducer.

Our approach is built around a simple principle: the vision-language model should provide structured visual evidence, and the temporal logic should be reduced explicitly. Instead of asking a model to answer every question from a global impression of the video, we parse the question, route the input, extract targeted evidence, and then apply operator-specific decision rules. This decomposition makes the system more robust to repeated events, short clips with dense actions, and long videos where only a few moments are relevant to a question.

%% file: sec/2_method.tex
\section{Method}

\subsection{Overview}

\begin{figure*}[t]
\centering
\includegraphics[width=0.98\textwidth]{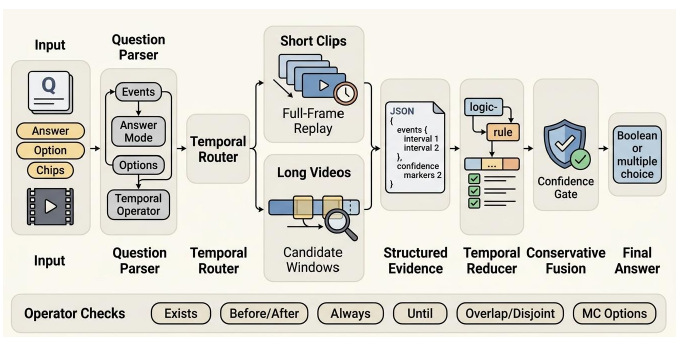}
\caption{Temporal Evidence Routing. The pipeline parses the question, routes the video input, obtains structured event evidence from targeted visual context, and applies temporal rules before final fusion.}
\label{fig:pipeline}
\end{figure*}

Figure~\ref{fig:pipeline} shows the overall Temporal Evidence Routing pipeline. The system has six stages: question parsing, event canonicalization, duration-aware visual evidence construction, structured VLM evidence generation, temporal micro-program verification, and temporal reduction with conservative fusion.

\subsection{Question Parsing and Operator Routing}

For each question, we identify the answer mode, event phrases, candidate options, and temporal operator. Multiple-choice questions are decomposed into independent option checks instead of being treated as a single global classification problem. This is important because an option may be visually present but still fail the requested temporal relation.

The parser maps each question into a compact program. A Boolean before question becomes a relation between two event variables, while a multiple-choice always-before question becomes four option-specific programs that share the same temporal condition. The parsed operator also controls the evidence budget. Existence questions can be answered from coarse evidence, while always, until, and overlap questions require more careful occurrence lists and counterexample checks.

\subsection{Event Canonicalization}

The parsed event phrases are normalized before visual reasoning. We separate action predicates from object arguments, map surface forms such as \emph{holding}, \emph{taking}, \emph{putting}, \emph{opening}, and \emph{closing} into transition-aware predicates, and attach each phrase to its role in the temporal program. This step prevents the reducer from treating semantically similar but temporally different states as equivalent.

For ambiguous verbs, the system builds strict visual definitions. A \emph{take} event requires a transition into possession, a \emph{put} event requires release to a support or location, and an \emph{open} or \emph{close} event requires a visible state transition rather than merely seeing the object. These definitions are used both in the VLM prompt and in later confidence checks.

\subsection{Visual Evidence Construction}

The visual input is selected according to video duration and temporal difficulty.

\begin{itemize}
  \item \textbf{Short clips.} For clips under a few seconds, we decode the clip into an ordered frame sequence and provide the frames as a slow replay. The frame index becomes the temporal axis, which avoids losing transient actions through sparse sampling.
  \item \textbf{Long videos.} For longer videos, we build event-focused candidate windows. A frame bank is sampled over the full video, visual-text similarity retrieves candidate moments for each event phrase, and neighboring frames are merged into temporal windows. The VLM only receives the windows that are relevant to the parsed events and options.
\end{itemize}

This routing reduces the burden on the VLM. Short clips preserve dense temporal changes, while long videos avoid asking the model to search an entire procedure for a small set of events. In the long-video retrieval branch, we use OpenAI CLIP ViT-B/16 \cite{clip} and {google/siglip-base-patch16-224} \cite{sclip} as local candidate-window selectors, followed by a stronger multimodal model for visual verification.

\subsection{Structured Visual Evidence}

The VLM is prompted to classify and describe events rather than directly answer the input question. Each event is classified as present, not visible, or uncertain, with frame or window-level evidence. For repeated actions, the model must list separate occurrences. For ambiguous action verbs such as \emph{hold}, \emph{take}, \emph{put}, \emph{open}, and \emph{close}, the prompt uses strict visual definitions so that result states and direct transitions are not conflated.

The structured output contains:
\begin{itemize}
  \item event status and confidence,
  \item one or more occurrence intervals or ordered frame ranges,
  \item short visual evidence for each event,
  \item near-miss evidence when an event is visually similar but not exact.
\end{itemize}

This evidence-first prompting follows the observation that modern VLMs are strong visual recognizers but can be unstable when asked to combine perception, event localization, and temporal logic in one step. We therefore use the multimodal model primarily as an evidence generator.

\subsection{Temporal Micro-Program Verification}

For dense short clips, the system adds a second verifier after structured evidence extraction. The verifier constructs a compact set of action instances from the ordered frame replay: each instance stores the canonical predicate, visible object arguments, frame range, and confidence. It then executes the parsed temporal program directly over this instance list.

This stage is especially useful for state-transition actions. Global VLM evidence may say that a person is holding an object, but the temporal program may require whether the person \emph{took} it before another event or merely held it afterward. The micro-program verifier therefore distinguishes persistent states from transition events and rejects near misses. Its output is not used as an unconditional replacement; it becomes a candidate answer with an explicit confidence and a reason code.

\subsection{Temporal Reduction}

After evidence extraction, a deterministic reducer applies the operator semantics. The reducer uses the extracted occurrence order and intervals, not the wording order of the question. The core rules are:

\begin{itemize}
  \item \textbf{Before / after:} compare the ordered occurrences of the two events.
  \item \textbf{Always before / after:} verify that the relation holds across all supported occurrences and reject visible counterexamples.
  \item \textbf{Overlap / disjoint:} compare event intervals and their intersections.
  \item \textbf{Until:} identify the maintained event and the boundary event, then check whether the evidence supports the boundary relation.
  \item \textbf{Multiple choice:} evaluate each option independently and choose the option that satisfies the temporal program.
\end{itemize}

If the evidence is missing or inconsistent, the reducer marks the case as unresolved instead of forcing a low-confidence answer. This is especially important for always and overlap questions, where a single missed occurrence can reverse the answer.

\subsection{Conservative Fusion}

The final stage combines the base prediction with candidate answers from the routed evidence pipeline. A candidate replacement is accepted only when the event evidence, temporal reducer, answer confidence, and micro-program reason code are consistent. Otherwise, the system keeps the base prediction. This conservative policy is designed to improve high-confidence temporal errors without introducing many noisy changes.

%% file: sec/3_exp.tex
\section{Experiments and Result}

\subsection{Setup}

The system is inference-only and does not fine-tune model weights. The main multimodal model used in the final run is {gemini-3.1-pro-preview} ~\cite{gemini}. Structured evidence extraction is generated in JSON mode with temperature $0.0$, {max\_output\_tokens}=4096, and {thinking\_level=MEDIUM}. The final answer reducer also uses JSON mode with temperature $0.0$, {max\_output\_tokens}=1024, and {thinking\_level=LOW}. For video input, clips shorter than $2.0$ seconds are decoded into all frames and passed as a chronological slow replay; clips from $2.0$ to $4.0$ seconds are represented by 12 uniformly sampled frames; longer videos use operator-dependent video sampling metadata.

The local retrieval branch first builds a frame bank for each long video, with 8 fps for videos under 10 seconds, 1-second stride for 10--60 second videos, and 2-second stride for videos at least 60 seconds long. Frames are resized to a maximum width of 640 pixels. We then encode every frame and each parsed event phrase with two retrieval backbones: OpenAI CLIP ViT-B/16 ~\cite{clip} and {google/siglip-base-patch16-224} ~\cite{sclip}. Image and text features are L2-normalized, and frame relevance is scored by the dot product between image and text embeddings. Each backbone returns the top 32 frames for each event. The two rankings are fused by reciprocal-rank fusion with $k=60$, and the top 8 fused frames are expanded into candidate windows. Each selected timestamp is expanded by $\pm 6$ seconds, windows within 2 seconds are merged, long windows are capped at 20 seconds, and at most 16 windows are passed to the VLM. For always and until questions, three additional coverage windows centered at 15\%, 50\%, and 85\% of the video are added. Confidence values are clipped to $[0,1]$, and unresolved evidence or evidence below the $0.55$ confidence threshold is handled conservatively.

\subsection{Official Test Result}

Table~\ref{tab:result} reports the final official test metric. We include only the final verified score.

\input{tables/main_test_results}

\subsection{Ablations}

Table~\ref{tab:ablations} compares direct Gemini prompting with the final Temporal Evidence Routing system. The baseline asks the multimodal model to answer from the video context, while our final system decomposes the task into evidence extraction, temporal micro-program verification, deterministic reduction, and conservative fusion. The comparison isolates the value of explicit temporal reasoning around the same multimodal backbone.

\input{tables/ablation_summary}

\subsection{Observations}

The most important empirical observation is that temporal errors are concentrated in operator-specific patterns rather than uniformly distributed across the test set. Direct full-video prompting is often sufficient for simple existence and coarse ordering questions, but it is less reliable for repeated events, all-occurrence constraints, and overlap relations. Separating event evidence from temporal reduction improves control over these cases.

Short clips benefit from full-frame replay because the temporal signal is compressed into very few seconds. Longer videos benefit from event-focused retrieval because the relevant events may occupy only a small part of the video. Multiple-choice questions require independent option checks; otherwise, the model may choose an option because it is visible or semantically plausible even when it violates the temporal relation. The large gap over direct Gemini prompting shows that the main bottleneck is not raw visual recognition alone, but converting visual observations into the exact temporal relation requested by the question.

%% file: tables/main_test_results.tex
\begin{table}[t]
\centering
\caption{Official test result.}
\label{tab:result}
\small
\setlength{\tabcolsep}{6pt}
\begin{tabular}{lc}
\toprule
System & AvgAcc \\
\midrule
Temporal Evidence Routing & \textbf{81.8} \\
\bottomrule
\end{tabular}
\end{table}

%% file: tables/ablation_summary.tex
\begin{table}[t]
\centering
\caption{Gemini-only baseline versus the final system.}
\label{tab:ablations}
\footnotesize
\setlength{\tabcolsep}{4pt}
\begin{tabular}{@{}p{0.66\linewidth}cc@{}}
\toprule
Configuration & AvgAcc & Gain \\
\midrule
Gemini-only direct prompting & 62.50 & -- \\
Temporal Evidence Routing & \textbf{81.8} & \textbf{+19.30} \\
\bottomrule
\end{tabular}
\end{table}

%% file: sec/4_conclusion.tex
\section{Conclusion}

We presented Temporal Evidence Routing, an inference-only approach for temporal-logic video QA. The system decomposes the task into question parsing, routed visual evidence construction, structured event evidence generation, deterministic temporal reduction, and conservative answer fusion. This design reduces reliance on global video impressions and makes the reasoning path more aligned with the temporal operators in TimeLogicQA. The final system achieves an official AvgAcc of 81.8 on the test evaluation.